\pdfoutput=1

\documentclass[11pt]{article}

\usepackage{acl} 

\usepackage{times}
\usepackage{latexsym}
\usepackage{graphicx}
\usepackage[T1]{fontenc}

\usepackage{lipsum}

\usepackage{pifont}       
\usepackage{bbding}       
\usepackage{fontawesome}  
\usepackage{bm}
\usepackage{adjustbox}
\usepackage{xcolor,colortbl}
\definecolor{Gray}{gray}{0.94}

\usepackage[utf8]{inputenc}
\usepackage{multirow}
\usepackage{microtype}
\usepackage{inconsolata}
\usepackage{booktabs}
\usepackage{xcolor, colortbl}	

\usepackage{hyperref} 

\usepackage{enumitem}

\usepackage{amssymb}

\definecolor{purple1}{RGB}{96,28,182} 
\definecolor{white1}{RGB}{255,255,255} 
\usepackage[listings]{tcolorbox}
\tcbuselibrary{listings,theorems,breakable}
\newtcolorbox{mybox}{colback=white!5!white,colframe=black!75!black, left=.05in, right=.05in}

\newtcbtheorem[number within=section]{exmp}{Example}%
{breakable,colback=white!5!white,colframe=purple1,fonttitle=\bfseries, left=.02in, right=.02in,bottom=.02in, top=.02in}{exmp}

\newtcbtheorem[number within=section]{trainprompt}{Prompt}%
{breakable,colback=white!5!white,colframe=purple1,fonttitle=\bfseries}{trainprompt}

\definecolor{darkgreen}{RGB}{50,100,0}
\definecolor{darkred}{RGB}{200, 0, 0}
\definecolor{lightred}{RGB}{250, 200, 200}
\definecolor{lightblue}{RGB}{208,190,248}
\newcommand{\blue}{\cellcolor{lightblue}}

\usepackage{soul} 

\soulregister{\textasciigrave}7

\usepackage{subfigure} 
\usepackage{makecell} 
\usepackage{multicol} 
\usepackage{float} 

\begin{document}


\title{MuMath-Code: Combining Tool-Use Large Language Models with Multi-perspective Data Augmentation for Mathematical Reasoning} 

\author{ 
Shuo Yin\textsuperscript{12}\footnotemark[2]\footnotemark[4], 
Weihao You\textsuperscript{1}\footnotemark[2], 
Zhilong Ji\textsuperscript{1}\footnotemark[1], 
Guoqiang Zhong\textsuperscript{2}, 
Jinfeng Bai\textsuperscript{1} \\
  \textsuperscript{1}Tomorrow Advancing Life \\ 
  \textsuperscript{2}College of Computer Science and Technology, Ocean University of China \\
  \texttt{shuoyinn@foxmail.com, youweihao@tal.com}, \\ \texttt{jizhilong@tal.com, gqzhong@ouc.edu.cn, jfbai.bit@gmail.com}
}

\maketitle

\renewcommand{\thefootnote}{\fnsymbol{footnote}} 
\footnotetext[2]{Equal contribution.} 
\footnotetext[4]{Work done while the author was interning at TAL.} 
\footnotetext[1]{Corresponding author.} 







\begin{abstract}
The tool-use Large Language Models (LLMs) that integrate with external Python interpreters have significantly enhanced mathematical reasoning capabilities for open-source LLMs, while tool-free methods chose another track: augmenting math reasoning data. However, a great method to integrate the above two research paths and combine their advantages remains to be explored. In this work, we firstly include new math questions via \textbf{mu}lti-perspective data augmenting methods and then synthesize \textbf{code}-nested solutions to them. The open LLMs (i.e., Llama-2) are finetuned on the augmented dataset to get the resulting models, \textbf{MuMath-Code} ($\mu$-Math-Code). During the inference phase, our MuMath-Code generates code and interacts with the external python interpreter to get the execution results. Therefore, MuMath-Code leverages the advantages of both the external tool and data augmentation. To fully leverage the advantages of our augmented data, we propose a two-stage training strategy: In Stage-1, we finetune Llama-2 on pure CoT data to get an intermediate model, which then is trained on the code-nested data in Stage-2 to get the resulting MuMath-Code.
Our MuMath-Code-7B achieves 83.8 on GSM8K and 52.4 on MATH, while MuMath-Code-70B model achieves new state-of-the-art performance among open methods---achieving 90.7\% on GSM8K and 55.1\% on MATH. Extensive experiments validate the combination of tool use and data augmentation, as well as our two-stage training strategy.
We release the proposed dataset along with the associated code for public use.
\end{abstract}

\section{Introduction} 


In Natural Language Processing (NLP), Large Language Models (LLMs)~\citep{gpt2,gpt3,t5} especially the proprietary ones such as GPT-4~\citep{openaichatgpt} and Claud-3~\citep{claude3} have \begin{figure}[t]
    \centering
    \includegraphics[width=.81\linewidth]{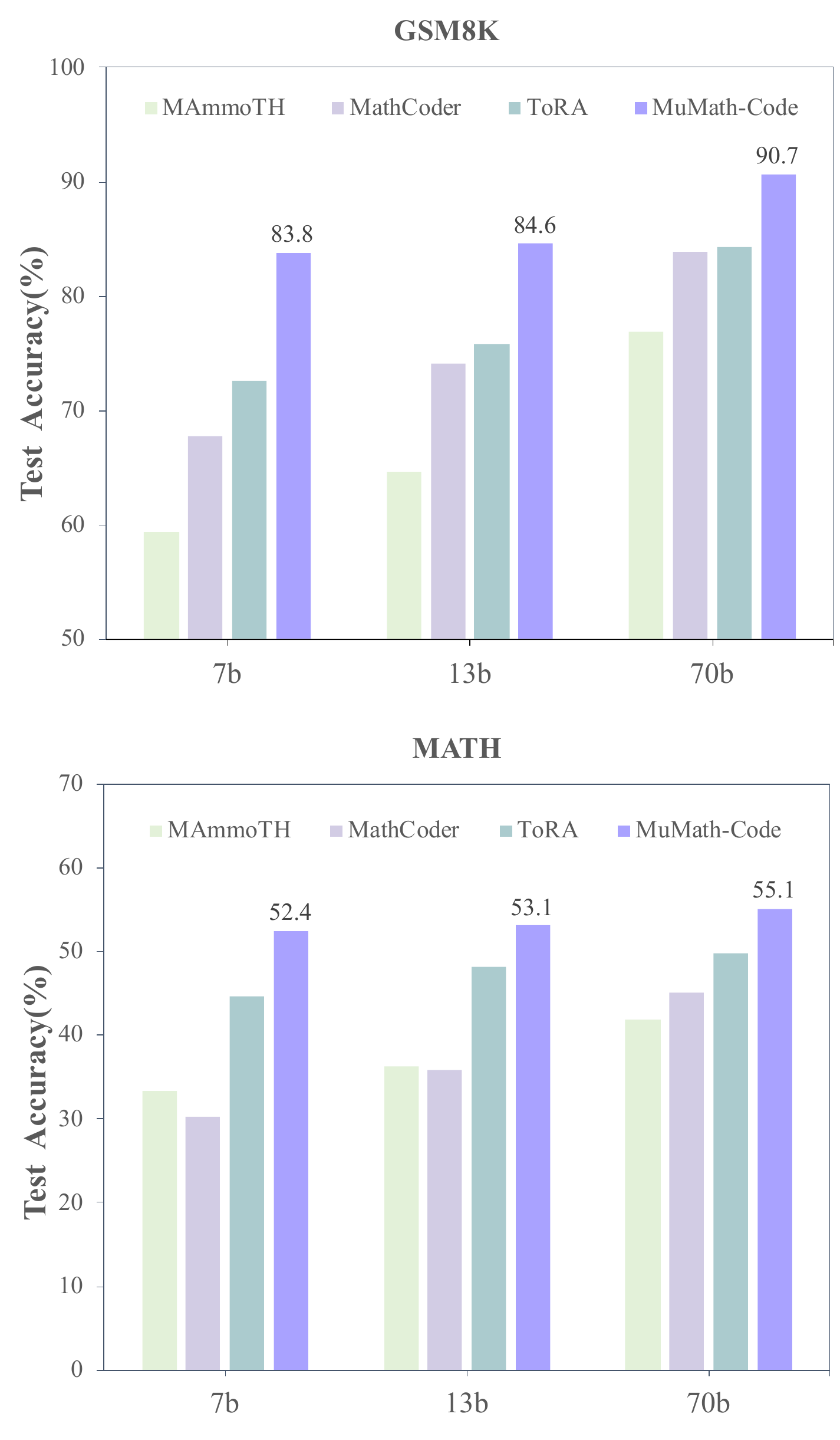}
    \caption{The comparison between our MuMath-Code and other state-of-the-art tool-use LLMs. MuMath-Code exhibits a substantial improvement in performance on both GSM8K~\citep{gsm8k} and MATH~\citep{MATH}, relative to the previous approaches.}
    \label{fig:head_figure}
\end{figure} demonstrated superiority in a variety of tasks, e.g., text classification~\citep{glue,bert,min-etal-2022-metaicl,pmlr-v202-jiang23k}, automated coding~\citep{chen2021codex,wizardcoder}, instructions following~\citep{longpre2023flan}, and math problem solving~\citep{chowdhery2022palm,minerva,anil2023palm,fu2023chainofthought}. Among these tasks, the capability to handle math problems stands as a typical and critical criterion for the evaluation of different LLMs. However, a significant performance disparity is observed between open-source LLMs, for instance, LLaMA~\citep{llama,llama2}, and their proprietary counterparts, when it comes to mathematical reasoning ability.

In recent years, many scholarly publications have been directed towards improving the mathematical proficiency of LLMs, which can be categorized into two distinct research trajectories: those that purely rely on natural language reasoning and those that incorporate external tools. The former methods are tool-free, mainly depends on data augmentation to enhance the models' mathematical reasoning capability, while the second trajectory (namely tool-use LLMs) are often coupled with external Python interpreters. From the perspective of knowledge distillation~\citep{huang2022large,li2022explanations,magister2023teaching,ho2023large,fu2023specializing,shridhar-etal-2023-distilling}, both mainstream approaches transfer math reasoning abilities from the powerful teacher models (for instance, GPT-4) to the inferior open foundation models.

The tool-free methods synthesize a large number of new math problems and corresponding solutions, taking the original training math QA pairs as the initial data seeds. Scaling law theoretically provides the basis for the ongoing improvement of LLMs' performance by constantly incorporating new training data. Representative approaches are RFT~\citep{rft}, MetaMath~\citep{yu2023metamath}, WizardMath~\citep{luo2023wizardmath}, MuggleMath~\citep{mugglemath}, MuMath~\citep{you2024mumath}, etc.
As for the second trajectory, code executors substantially supplant LLMs in particularly challenging computational and logical tasks, thereby alleviating the problem-solving burden on them. This tool-use category is exemplified by PAL~\citep{gao2023pal}, PoT~\citep{chen2023program}, MAmmoTH~\citep{mammoth}, ToRA~\citep{tora} and MathCoder~\citep{mathcoder}. 


Although the aforementioned research paths have been individually successful, to date, few methods have been developed that amalgamate their respective advantages. In this paper, we propose a novel method that integrates tool usage with data augmentation to synthesize a large amount of \textbf{mu}lti-perspective \textbf{math}ematical questions and solutions (we employ the augmenting methods introduced in a previous work MuMath~\citep{you2024mumath}). Specifically, we utilize proprietary LLMs (like GPT-4) to generate Python \textbf{code} while synthesizing new solutions to math problems, and then fine-tune the open-source models (e.g., LLaMA) on the augmented dataset. The resulting model, \textbf{MuMath-Code}, is thus equipped with the ability to write code for math problem solving. During the inference phase, our MuMath-Code can generates both CoT~\citep{cot} reasoning texts and Python code blocks. These code blocks are then extracted and executed by an external Python interpreter, and the execution results are returned to MuMath-Code for subsequent rounds of CoT reasoning or code generation until the final result is obtained or the maximum number of execution rounds is reached.

The multi-perspective mathematical question set comprises questions augmented via rephrasing~\citep{yu2023metamath}, alteration~\citep{mugglemath,you2024mumath}, FOBAR~\citep{fobar}, BF-Trans~\citep{you2024mumath}, besides those from the original training sets.
Regarding the solutions nested with Python code, we leverage a general pattern like the ones used in ToRA~\citep{tora} and MathCoder~\citep{mathcoder}: CoT-PoT interleaving. However, we propose prefix CoT, code debugging and pseudo-answer guidance filtering to improve the consistency and quality of our augmented solutions. The prefix CoT is a thoughtful analysis in pure natural language before code generation, making the LLMs consider this analysis while generating all the subsequent content, which thus are helpful for the models to learn the whole solution. Besides, we prompt GPT-4 to debug and correct the inexecutable code when requesting the solutions, and we keep the faulty code since this process of verification and correction can help boost the models' coding proficiency. Furthermore, for those synthesized questions via alteration, which lack ground truth answers as filtering guidance, we choose the majority-voting answers as the pseudo-answers. This process can increase the correctness of the generated solutions and thus improve the data quality generally. We name the proposed dataset as \textbf{MuMath-Code-Data} and denote it as $\mathcal{D}_{\mu\text{-}code}$. 

Moreover, previous tool-use LLMs for math are derived by directly finetuning on code-nested data, which thus fail to fully harness the intrinsic natural language reasoning capability of the LLMs themselves. Different from the other tool-use methods, we design a two-stage training strategy to better combine the advantages of data augmentation and external code execution. The first stage is to enhance the models' pure language mathematical reasoning, where the largest (751K) dataset proposed in MuMath (here called \textbf{MuMath-Data} and denoted as $\mathcal{D}_\mu$) is utilized to finetune LLaMA, and get an intermediate model, \textbf{MuMath}. In the second stage, we continue finetuning MuMath on MuMath-Code-Data to equip the model with the ability to write code for solving math problems. The resulting model, \textbf{MuMath-Code}, is thus can be prompted to leverage the Python interpreter to execute its generated code for securing the desirable outputs at inference time. 

Our contributions are summarized as follows:
\begin{itemize}[topsep=0pt, partopsep=0pt]
\setlength{\itemsep}{0pt}


\item  We construct a multi-perspective augmentation dataset with code-nested solutions for math problem solving, called MuMath-Code-Data. 
\item We design a two-stage training strategy to equip the open LLMs with pure language reasoning and math related code generation capabilities, respectively.
\item The obtained model, MuMath-Code, achieves new
state-of-the-art performance among open LLMs across the in-domain math reasoning datasets as well as the out-of-domain ones. MuMath-Code-7B have 83.8 on GSM8K and 52.4 on MATH, while MuMath-Code-70B has achieved 90.7\% on GSM8K and 55.1\% on MATH.
\end{itemize} 

\section{Related Work} 

\subsection{Tool-Free LLMs for Math} 

Rejection Sampling-based Fine-Tuning (RFT,~\citealp{rft}) only augments the solutions via rejection sampling to collect a variety of different reasoning paths. Since RFT does not introduce new math questions, the diversity of the augmented dataset is quite low, which limits the performance improvement of the finetuned models.
With the aim of incorporating a broader spectrum of questions, MetaMath~\citep{yu2023metamath} employs rephrasing, Self-Verification (SV,~\citealp{sv}) and FOBAR~\citep{fobar} to generate new questions. Ideally speaking, like the original questions, there are also ground truth answers for filtering solutions to these augmented questions. 
To bring in more diverse data, WizardMath~\citep{wizardlm,luo2023wizardmath} and MuggleMath~\citep{mugglemath} choose to create totally new questions via evolution or directional modification (changing numbers, adding conditions, increasing complexity, etc.) based on the seed questions. These altered questions have no ground truth answers, thus lacking a criterion to filter their corresponding synthesized solutions. 

Furthermore, MuMath~\citep{you2024mumath} leverages some of the aforementioned methods, and additionally proposes BF-Trans and expression replacement (etc.) to perform comprehensive augmentation, thus constructing a multi-perspective math question set with much greater diversity. For improving data quality, majority sampling serves as the filtering rule for the synthesized solutions to those new questions without deterministically known answers. Instead of solution filtering, a contemporary work, Xwin-Math~\citep{xwin_math}, employs verification with solution requesting during question synthesis, thereby improving the solvability of the questions and the correctness of the answers. Since there is no restriction on the direction of question modification, Xwin-Math theoretically offers a wider variety of diverse synthesized data. Balancing the efficacy and the ease of replication, in this paper the proposed MuMath-Code opts to employ the question augmentation from MuMath, although it is orthogonal to any other augmentation methods.

Nevertheless, as probabilistic models, LLMs inherently have limitations in logical reasoning and numerical computation. Thus, to improve the accuracy of mathematical problem-solving while relying solely on the capabilities of LLMs necessitates the utilization of a substantially larger dataset compared to tool-use methods.

\subsection{Tool-Use LLMs for Math} 
Another research trajectory highlights the synergy between LLMs and external tools. Pioneering efforts along this include the Program-aided Language model (PAL,~\citealp{gao2023pal}) and Program of Thought (PoT,~\citealp{chen2023program}). Moreover, MAmmoTH~\citep{mammoth} integrates both CoT and PoT in a coarse-grained fashion (each sample corresponds to only one of these two possible solution types), enabling flexible inference where the finetuned models may adopt different methods for different questions. Different from MAmmoTH, ToRA~\citep{tora} interleaves python code blocks and natural language reasoning parts over multiple turns for a same solution, which offers a more flexible combination of CoT and PoT. However, neither MAmmoTH nor ToRA employs query augmentation, thereby narrowing the range of math questions, which in effect, limits the problem-solving capabilities that can be acquired. Wang et al. propose a contemporaneous work with ToRA, MathCoder~\citep{mathcoder}, where each solution is also organized in an interleaved manner. Besides, they introduce interpolation problems to mitigate the disparity in difficulty level between GSM8K~\citep{gsm8k} and MATH~\citep{hendrycks2021measuring}. Hence, like our MuMath-Code, MathCoder is also an amalgamation of tool usage and math question augmentation, although the new questions it introduces are comparatively narrow in scope and limited in diversity.

Similar to ToRA and MathCoder, we also construct such solutions  that intertwine Python code with pure language reasoning text to adaptably combine LLMs with external code executing tools. However, we propose prefix CoT, code debugging, and pseudo-answer guidance filtering to further enrich the solutions and improve their correctness. Additionally, different from MathCoder, the question augmentation we utilize are multi-perspective, thus offering greater diversity and exposing the model to a broader scope of novel questions, thereby significantly enhancing the model's generalization capabilities.

\begin{figure*}[t]
    \centering
    \includegraphics[width=.98\linewidth]{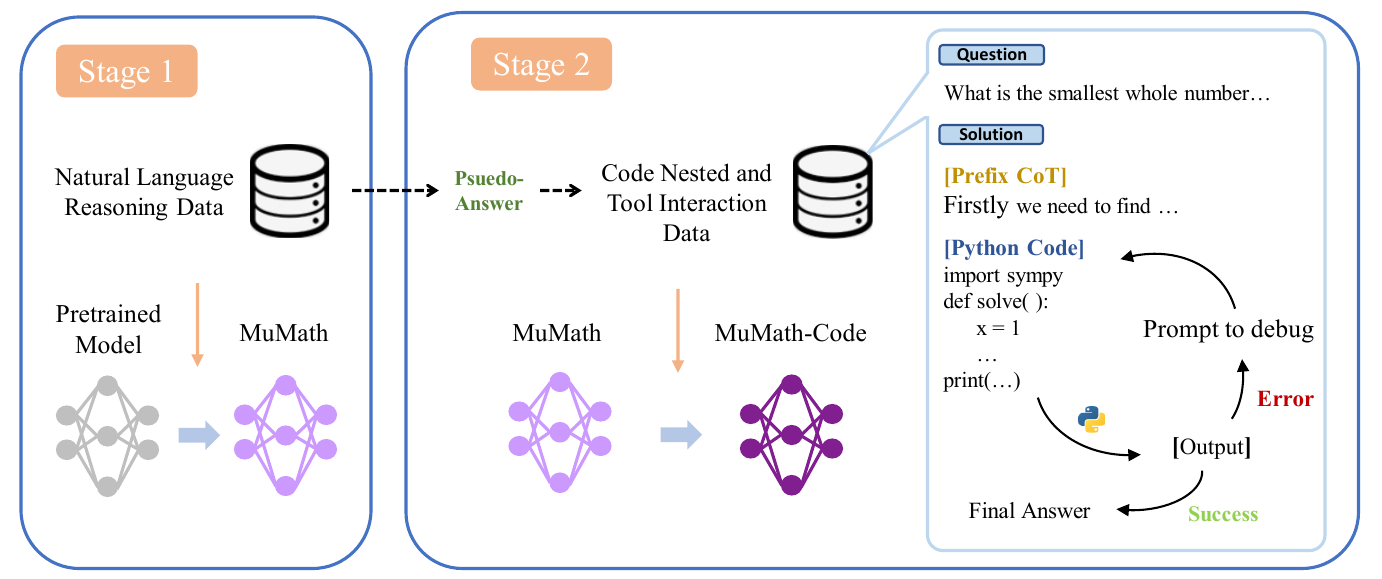}
    \caption{Illustration of our proposed method. The foundation model is first trained through an initial stage, resulting in an intermediary model that possesses more powerful math reasoning capability. This intermediary model is then further trained on the proposed dataset to learn code generation and tool interaction, leading to the final model, MuMath-Code.}
   \label{fig:methodology}
\end{figure*}

\section{Preliminaries}


We employ the augmented questions from MuMath~\cite{you2024mumath} and synthesize code-nested solutions to them. To help the models better learn such solutions with multi-turn code generation, code execution and pure natural language reasoning, we propose prefix CoT, code debugging, and pseudo-answer guidance filtering to augment the quality of the synthetic data, as well as a two-stage training strategy. Figure~\ref{fig:methodology} delineates the overall pipeline.






\subsection{MuMath Augmented Questions} 

The original questions from the training sets of GSM8K~\citep{gsm8k} and MATH~\cite{MATH} are taken as the seed question set $Q_{original}$. The question augmenting methods employed in MuMath are conducted on this seed set, which are briefly concluded as follows: 

\paragraph{(1) Rephrasing} Rewrite a text while keeping the original meaning unchanged. Based on the fact that a rephrased question holds the same meaning as the original one, the final answer of it should also be the same. We denote the rephrased question set as $Q_{rephrase}$. 

\paragraph{(2)  Question Alteration} There are five manners to alter the original questions, like changing numbers and adding more conditions, concluded in MuggleMath~\citep{mugglemath}. The resultant question set created via alteration is referred to as $Q_{alter} = Q_{alter1}\cup Q_{alter2}\cup Q_{alter3}\cup Q_{alter4}\cup Q_{alter5}$. Besides, Expression Replacement, proposed in MuMath, firstly get the expressions of the solution to an original question, then change the calculation operators within them. Based on the changed expressions, a new question is asked to generate. $Q_{replace}$ represents the question set produced by this augmentation technique. Note that $Q_{alter}$ and $Q_{replace}$ correspond no definitely correct answers due to modifications in the questions' intrinsic meanings.

\paragraph{(3) FOBAR} Following~\citet{fobar}, we mask a certain condition in an initial question by substituting it with ``X", and meanwhile give the answer to the original question as a new condition, thereby creating a reverse question that seeks to determine the value of the unknown X. $Q_{fobar}$ is utilized to mark the FOBAR question set. 

\paragraph{(4) BF-Trans} Backward-Forward Transformation (BF-Trans), proposed in MuMath, aims to construct such backward questions that can be answered through direct reasoning, bypassing the necessity of solving equations to find the unknown variables (thus resemble the data sampled from the original distribution). For a certain question-answer pair, BF-Trans firstly utilize FOBAR to transform the original question into a backward one; secondly, we rephrase the FOBAR question into a new form where the masked value is requested directly instead of employing an unknown variable X, resulting in a ``secondary forward'' question which we called BF-Trans question. The set of these BF-Trans questions is marked as $Q_{bf}$. 

To sum up, all the 10 aforementioned subsets (5 in $Q_{alter}$) constitute the resulting question set $\mathcal{Q} = Q_{original} \cup Q_{rephrase} \cup Q_{alter} \cup Q_{replace}\cup Q_{fobar} \cup Q_{bf}$.
Based on $\mathcal{Q}$, we generate 2 datasets called \textbf{MuMath-Data} and \textbf{MuMath-Code-Data}, emphasizing pure natural language mathematical reasoning and tool interaction via code generation, respectively.

\subsection{MuMath-Data} 
MuMath-Data (denoted as $\mathcal{D}_{\mu}$) is just the largest dataset from MuMath, which contains about 750K samples with pure CoT reasoning solutions to questions in $\mathcal{Q}$.

\paragraph{Majority Sampling} 
As is introduced in the paper of MuMath, for $Q_{alter}$ and $Q_{replace}$ whose each question has no reference answer, majority sampling is utilized to filter all the randomly generated solutions and only those solutions with the majority answers are kept. In other words, each majority answer serves as a pseudo-answer to the corresponding question.

\section{Methodology}

\subsection{MuMath-Code-Data} 
\label{subsection:mumath_code_data}


To facilitate the interaction with the python interpreter, we synthesize the code-nested solutions for the models to learn, each consisting of multi-turn code generation, code execution and pure natural language reasoning.

Specifically, for each question from $\mathcal{Q}$, we prompt proprietary LLMs to request solutions each with at least one block of code, which is then extracted and passed to the external interpreter for execution. Every execution result is appended to the preceding content, right after the corresponding code block. If the code execution fails, we append a prompt to actively debug, using all the previous content as a whole new prompt to request the corrected code, which we then extract and execute again. By iterating this process multiple times, we obtain a reasoning path comprising code, execution outcomes and natural language analysis. This reasoning path is similar to that of MathCoder~\citep{mathcoder} and ToRA~\citep{tora}, but the differences lie in the use of our proposed prefix CoT, code debugging, and pseudo-answer guidance filtering, which will be elaborated on in this section. We marked MuMath-Data-Code as $\mathcal{D}_{\mu\text{-}code}$.

\paragraph{Prefix CoT} 
We have observed that before generating code, a thorough pure natural language analysis is helpful for the models' performance. Therefore, we deliberately add a thoughtful CoT reasoning before code writing. The request prompt used is ``\textit{Analyze the question; list some knowledge points related to the question and beneficial for problem solving.}''.

\paragraph{Code Debugging} 
Several research studies have shown that the use of error correction and verification data can improve the mathematical reasoning capabilities of LLMs. Therefore, we introduce an error correction process for our augmented dataset. Specifically, while constructing a solution, if the generated code fails to execute, we append a prompt ``\textit{The code above has encountered a problem. Now point out its mistakes and then correct them.}'' for GPT-4 to debug the code and write new code until the executable code is obtained, or the maximum number of requests is reached. The failing code and error information are kept to equip the finetuned models with debugging ability, and thus enhance their coding proficiency for solving math problems.

\paragraph{Pseudo-Answer Guidance Filtering} 
In MuMath-Data, we employ majority sampling to filter solutions. This provides us with pseudo-answers for the augmented questions corresponding no reference answers, which can also be employed for MuMath-Code-Data to select solutions. This approach improve the correctness of the synthesized solutions, thereby leading to an enhancement in the overall quality of the augmented data.

To sum up, we mark the $i$-th CoT (pure natural language reasoning) part as $c_i$; the $i$-th python code part is marked as $p_i$, which always begins with \sethlcolor{lightgray}\hl{\textasciigrave{}\textasciigrave{}\textasciigrave{}python} and ends with \hl{\textasciigrave{}\textasciigrave{}\textasciigrave{}}; the $i$-th code execution output is denoted as $o_i$, beginning with \hl{\textasciigrave{}\textasciigrave{}\textasciigrave{}output} and ending with \hl{\textasciigrave{}\textasciigrave{}\textasciigrave{}}. To formalize, one resulting solution $s$ is defined as follows:

\begin{equation} 
\begin{split}
s 
&= \Big (\bigoplus\limits_{i=1}^{n-1} c_ip_io_i \Big )c_n \\ 
&= c_1p_1o_1c_2p_2o_2...c_{n-1}p_{n-1}o_{n-1}c_n, 
\end{split}
\end{equation} 
where $\bigoplus$ stands for the concatenation of all the turns, and $n$ is the number of CoT parts. See Appendix~\ref{appendix:examples} for an example.

\subsection{Two-Stage Training} 
\label{subsection:training}

\paragraph{Stage-1}
The first stage training is on MuMath-Data, where the models concentrate on learning the capability of pure CoT math reasoning. The learning target is as follows:

\begin{equation} 
\begin{split} 
&\mathcal{L}_1 = -\mathbb{E}_{q,s\sim \mathcal{D}_{\mu}} \bigg[\sum\limits_{t=1}^{l}
\log P\big(x_t | q, x_{<t}; \bm \theta \big)\bigg],
\end{split} 
\end{equation}
where the solution $s = (x_1, x_2, ..., x_l) $ contains $l$ tokens, and $\bm \theta$ is the parameter of MuMath-Code.

This training stage endow the models with a fairly strong mathematical reasoning capability, which can be seen as an preliminary task for the second stage learning.

\paragraph{Stage-2}
The second stage training is on MuMath-Code-Data, where the models concentrate on PoT-CoT interleaved data to learn how to interact with an external tool (i.e., the Python interpreter). We mask the loss of the outputs from the code execution, which should not be learned by the models. The learning target is:

\begin{equation} 
\begin{split} 
&\mathcal{L}_2 = \\
&-\mathbb{E}_{q,s\sim \mathcal{D}_{\mu\text{-}code}} \bigg[\sum\limits_{i=1}^{n}
\log P\big(c_ip_i|q, \bigoplus\limits_{j=1}^{i-1} c_jp_jo_j; \bm \theta \big)\bigg], 
\end{split}
\end{equation}
where $p_n = \varnothing$. The training process at Stage-2 is consistent with the inference, so we do not need to consider the issue of catastrophic forgetting (regarding the natural language reasoning in Stage-1). At inference time, after being given a mathematical problem, the finetuned model needs to generate code for problem solving, and then an external interpreter executes the code and returns the result for the model to continue generating. Therefore, Stage-2 training simulates the above inference process by masking out the losses of the execution outputs.

\section{Experiments} 



\subsection{Experimental Setup} 
\paragraph{Datasets} 
Our seed datasets for synthesis are the training sets of two popular math reasoning benchmarks: GSM8K~\cite{gsm8k} and MATH~\cite{MATH}. GSM8K contains elementary school math problems, comprising 7,473 training instances and 1,319 test instances; while MATH encompasses math competition problems at the high school level with 7,500 training samples and 5,000 for test. 

We take the MuMath~\cite{you2024mumath} dataset (750K) as our $\mathcal{D}_{\mu}$ for Stage-1 training, and the MuMath augmented question set $\mathcal{Q}$ are utilized to construct $\mathcal{D}_{\mu\text{-}code}$ for Stage-2; in $\mathcal{Q}$, we request 15 solutions for each question that originates from GSM8K and 30 for MATH-related ones, and then perform filtering to get 30K samples for each question subset, making 600K in total.

\paragraph{Implementation Details} 

Our study utilizes LLaMA-2 (7B, 13B and 70B)~\cite{llama2} and CodeLlama (7B, 13B, 34B, and 70B)~\citep{codellamapal} as the foundation models for full-parameter finetuning, corresponding to MuMath-Code-L and MuMath-Code-CL as the resulting models. We employ AdamW as the optimizer and a cosine learning rate scheduler with a 0.03 warmup ratio. Across all the models and both stages, we train 3 epochs with a 128 global batch size. All the models except for LLaMA-70B and CodeLlama-70B are trained using the Deepspeed framework, while those two 70B models are trained using Megatron for the sake of speed. The hardware we use are NVIDIA H800 GPUs.


\begin{table*}
\huge
\centering
\resizebox{.90\textwidth}{!}{
    \begin{tabular}{l ccccccc}
    \toprule[2.2pt]
    Model & GSM8K & MATH & GSM-Hard & SVAMP & TabMWP & ASDiv & MAWPS\\
    
    \midrule
    
    \multicolumn{8}{c}{\textit{colsed-source LLMs}}  \\ 

    \midrule 

    Claud-3 Opus~\citep{claude3} & 95.0 & 60.1 & - & - & - & - \\ 
    GPT-4 \cite{gpt4}  & 92.0 & 42.5 & 64.7 & 93.1 & 67.1 & 91.3 & 97.6 \\
    GPT-4 (PAL)  & 94.2 & 51.8 & 77.6 & 94.8 & 95.9 & 92.6 & 97.7 \\
    GPT-3.5 \citep{openaichatgpt} & 80.8 & 35.5 & 55.9 & 83.0 & 69.1 & 87.3 & 94.6 \\
    GPT-3.5 (PAL) & 78.6 & 38.7 & 67.6 & 77.8 & 79.9 & 81.0 & 89.4 \\
    %
    \midrule 

    \multicolumn{8}{c}{\textit{tool-free open LLMs}}  \\

    \midrule 
    
    \multicolumn{8}{c}{\textit{7B}} \vspace{2.5mm} \\
    
    LLaMA-2 \citep{llama2}  & 13.3 & 4.1 & 7.8 & 38.0 & 31.1 & 50.7 & 60.9 \\
    LLaMA-2 SFT \citep{llama2}  & 41.3 & 7.2 & 16.1 & 31.9 & 27.8 & 47.4 & 60.0 \\
    WizardMath \citep{luo2023wizardmath}  & 54.9 & 10.7 & 20.6 & 57.3 & 38.1 & 59.1 & 73.7 \\ 
    MetaMath \citep{yu2023metamath}  & 66.5 & 19.8 & - & - & - & - & - \\
    MuggleMath \citep{mugglemath}  & 68.4 & - & - & - & - & - & -  \\
    
    MuMath~\citep{you2024mumath}   & 70.9 & 22.0 & - & 76.8 & - & 93.6 & 87.3 \\

    \midrule
    
    \multicolumn{8}{c}{\textit{13B}} \vspace{2.5mm} \\
    
    LLaMA-2 \citep{llama2}  & 24.3 & 6.3 & 13.6 & 43.1 & 39.5 & 56.3 & 70.4 \\
    LLaMA-2 SFT \citep{llama2}  & 51.1 & 9.2 & 22.3 & 46.3 & 35.8 & 58.6 & 75.0 \\
    WizardMath \citep{luo2023wizardmath}  & 63.9 & 14.0 & 28.4 & 64.3 & 46.7 & 65.8 & 79.7 \\ 
    MetaMath \citep{yu2023metamath}  & 72.3 & 22.4 & - & - & - & - & -  \\
    MuggleMath \citep{mugglemath}  & 74 & - & - & - & - & - & -  \\
    MuMath~\citep{you2024mumath}   & 76.4 & 25.3 & - & - & - & - & - \\
    
    \midrule
    
    \multicolumn{8}{c}{\textit{70B}} \vspace{2.5mm} \\
    
    LLaMA-2 \citep{llama2}  & 57.8 & 14.4 & 36.0 & 73.6 & 57.5 & 76.0 & 92.4 \\
    LLaMA-2 SFT \citep{llama2}  & 69.3 & 14.9 & 39.0 & 64.0 & 53.0 & 71.3 & 84.8 \\
    WizardMath \citep{luo2023wizardmath}  & 81.6 & 22.7 & 50.3 & 80.0 & 49.8 & 76.2 & 86.2 \\ 
    MetaMath\citep{yu2023metamath}   & 82.3 & 26.6 & - & - & - & - & - \\
    MuggleMath \citep{mugglemath}  & 82.3 & - & - & - & - & - & -  \\
    MuMath~\citep{you2024mumath}   & 84.5 & 32.2 & - & 87.6 & - & 96.6 & 92.0\\
    
    \midrule

    \multicolumn{8}{c}{\textit{tool-use open LLMs}}  \\

    \midrule 
    \multicolumn{8}{c}{\textit{7B}} \vspace{2.5mm} \\

    MAmmoTH~\citep{mammoth}  & 53.6 & 31.5 & - & 67.7 & - & - & - \\
    MAmmoTH-Coder  & 59.4 & 33.4 & - & 71.4 & - & - & - \\
    CodeLLama (PAL)~\citep{codellamapal} & 34.0 & 16.6 & 33.6 & 59.0 & 47.3 & 61.4 & 79.6\\
    MathCoder-L \citep{mathcoder}  & 64.2 & 23.3 & - & 71.5 & - & - & - \\
    MathCoder-CL \citep{mathcoder}  & 67.8 & 30.2 & - & 70.7 & - & - & - \\
    ToRA~\citep{tora}  & 68.8 & 40.1 & 54.6 & 68.2 & 42.4 & 73.9 & 88.8 \\ 
    ToRA-Code~\citep{tora} & 72.6 & 44.6 & 56.0 & 70.4 & 51.6 & 78.7 & 91.3 \\ 
    \blue{\textbf{MuMath-Code-L}}   & \blue{\textbf{83.8}}& \blue{\underline{48.8}} & \blue{\underline{70.5}} & \blue{\underline{87.6}} & \blue{\underline{65.6}} & \blue{\underline{86.2}} & \blue{\underline{94.7}}  \\ 
    
    \blue{\textbf{MuMath-Code-CL}}   & \blue{\underline{82.6}}& \blue{\textbf{52.4}} & \blue{\textbf{70.6}} & \blue{\textbf{88.1}} & \blue{\textbf{66.9}} & \blue{\textbf{87.4}} & \blue{\textbf{95.3}}  \\ 
    
    \midrule
    
    \multicolumn{8}{c}{\textit{13B}} \vspace{2.5mm} \\

    MAmmoTH~\citep{mammoth}  & 62.0 & 34.2 & - & 72.4 & - & - & -\\
    MAmmoTH-Coder\citep{mammoth} & 64.7 & 36.3 & - & 73.7 & - & - & -\\ 
    CodeLlama (PAL)~\citep{codellamapal} & 39.9 & 19.9 & 39.0 & 62.4 & 59.5 & 65.3 & 86.0 \\
    MathCoder-L \citep{mathcoder}  & 72.6 & 29.9 & - & 76.9 & - & - & - \\ 
    MathCoder-CL \citep{mathcoder}  & 74.1 & 35.9 & - & 78.0 & - & - & - \\ 
    
    ToRA \citep{tora}  & 72.7 & 43.0 & 57.3 & 72.9 & 47.2 & 77.2 & 91.3 \\ 
    ToRA-Code~\citep{tora} & 75.8 & 48.1 & 60.5 & 75.7 & \underline{65.4} & 81.4 & 92.5 \\ 
    \blue{\textbf{MuMath-Code-L}}   & \blue{\underline{84.3}}& \blue{\underline{49.9}} & \blue{\underline{70.6}} & \blue{\textbf{87.9}} & \blue{64.9} & \blue{\textbf{86.4}} & \blue{\underline{94.9}} \\ 
    
    \blue{\textbf{MuMath-Code-CL}}   & \blue{\textbf{84.6}}& \blue{\textbf{53.1}} & \blue{\textbf{70.8}} & \blue{\underline{86.8}} & \blue{\textbf{67.2}} & \blue{\underline{85.2}} & \blue{\textbf{95}} \\
    
    \midrule 

    \multicolumn{8}{c}{\textit{34B}} \vspace{2.5mm} \\

    CodeLLaMa (PAL) \citep{codellamapal} & 53.3 & 23.9 & 49.4 & 71.0 & 63.1 & 72.4 & 91.5 \\ 
    MAmmoTH-Coder~\citep{mammoth} & 72.7 & 43.6 & - & 84.3 & - & - & - \\

    MathCoder-CL~\citep{mathcoder} & 81.7 & 45.2 & - & 82.5 & - & - & - \\
    
    ToRA \citep{tora}  & 80.7 & 50.8 & 63.7 & 80.5 & 70.5 & 84.2 & \textbf{93.3}  \\ 
    
    \blue{\textbf{MuMath-Code-CL}}   & \blue{\textbf{87.6}}& \blue{\textbf{55.0}} & \blue{\textbf{68.8}} & \blue{\textbf{91.4}} & \blue{\textbf{74.9}} & \blue{\textbf{87.9}} & \blue{\underline{92.9}} \\

    \midrule 
    
    \multicolumn{8}{c}{\textit{70B}} \vspace{2.5mm} \\

    LLaMA-2 (PAL) & 55.2 & 18.3 & 50.0 & 74.6 & 59.5 & 71.9 & 92.8 \\
    
    MAmmoTH \citep{mammoth}  & 76.9 & 41.8 & - & 82.4 & - & - & - \\
    
    MathCoder-L \citep{mathcoder}  & 83.9 & 45.1 & - & 84.9 & - & - & - \\
    ToRA~\citep{tora}  & 84.3 & 49.7 & 67.2 & 82.7 & 74.0 & 86.8 & 93.8 \\ 
    \blue{\textbf{MuMath-Code-L}}   & \blue{\textbf{90.7}}& \blue{\underline{52.8}} & \blue{\underline{68.6}} & \blue{\textbf{93}} & \blue{\underline{74}} & \blue{\textbf{88.4}} & \blue{\textbf{95.4}} \\ 
    \blue{\textbf{MuMath-Code-CL}}   & \blue{\underline{89.5}}& \blue{\textbf{55.1}} & \blue{\textbf{70.1}} & \blue{\underline{92.9}} & \blue{\textbf{77.4}} & \blue{\underline{87.9}} & \blue{\underline{94.7}} \\
    \bottomrule[2.2pt]

    \end{tabular}}
    \caption{Comparison of the state-of-the-art methods on various datasets. For the \textit{tool-use open LLMs}, the best results are bolded and the second best underlined among the same scale models tested on the same datasets. }
    \label{table:comparison}
\end{table*}


\subsection{Comparison Results} 

As shown in Table~\ref{table:comparison}, the comparison experiment of our models with the current state-of-the-art demonstrates that our approach consistently achieves superior performance across all scales of open-source models on all the datasets. Notably, our MuMath-Code-L 7B model has attained a test accuracy of 83.8 on the GSM8K, and MuMath-Code-CL 7B has reached a score of 52.4 on MATH. These outcomes surpass many 70B open-source baselines and even some proprietary LLMs. Additionally, our MuMath-Code-CL 34B and 70B achieve 55.0+ on MATH, two impressive results considering that they are accomplished by leveraging data augmentation techniques based on the original training set without the incorporation of extensive additional mathematical corpora for pre-training.


There are some noteworthy findings from the experimental statistics presented in the table, such as the performance of MuMath-Code-CL 13B on MATH, registering at 53.1, which is only marginally higher than that of MuMath-Code-CL 7B, which stands at 52.4. Moreover, the MuMath-Code-CL 34B's performance on MATH, scoring at 55.0, is very close to that of the MuMath-Code-CL 70B, which records a score of 55.1. We speculate that this may be attributed to the phenomenon where, beyond a certain threshold of data volume, the advantages conferred by increased model size may be diminished or even offset by the benefits derived from the expanded dataset. Additionally, variations in the training frameworks may also contribute to the observed discrepancy between the performances of MuMath-Code-CL 34B and 70B.

\subsection{Effectiveness of the Two-Stage Training Strategy} 

\begin{table}[h]
\huge
\centering
\renewcommand\arraystretch{1.5} 
\resizebox{.48\textwidth}{!}{
    
    \begin{tabular}{cccccc}
    \toprule[2.2pt] 
    \multirow{2}{*}{\makecell{ Inference }} & \multirow{2}{*}{Training Strategy} & \multicolumn{2}{c}{LLaMA} & \multicolumn{2}{c}{CodeLlama} \\ 
     & & GSM8K & MATH & GSM8K & MATH \\ 

    \midrule 
     
     \multirow{2}{*}{Tool free} 
     & $\mathcal{D}_{meta}$ & 66.5 & 19.8 & - & - \\ 
     & $\mathcal{D}_{xwin} $ & 66.6 & 17.4 & - & - \\ 
     
     \midrule 
     
     \multirow{5}{*}{Tool use} 
     & $\mathcal{D}_{\mu\text{-}code}$ & 81.2 & 46.2 & 81 & 49.8 \\ 
     & $\mathcal{D}_\mu + \mathcal{D}_{\mu\text{-}code} $ & 82.7 & 47.1 & 81.3 & 49.1 \\ 
     & $\mathcal{D}_{meta} \rightarrow \mathcal{D}_{\mu\text{-}code} $ & 82.3 & 47.4 & - & - \\ 
     & $\mathcal{D}_{xwin} \rightarrow \mathcal{D}_{\mu\text{-}code} $ & 82.0 & 47.2 & - & - \\ 
     & $\mathcal{D}_\mu \rightarrow \mathcal{D}_{\mu\text{-}code} $ & 83.8 & 48.8 & 82.6 & 52.4 \\ 

    \bottomrule[2.2pt] 
    \end{tabular}
    }
    \caption{A two-stage training strategy improves the models' performance, as opposed to a single-stage training. }
    \label{table:two_stage}
\end{table}

MuMath-Code is derived from a two-stage training process that enhances the model's pure natural language reasoning capabilities and the ability to generate code and interact with external tools. In this section, we validate the efficacy of this bifurcated training strategy. Unless otherwise specified, all ablation experiments presented in this paper are conducted on 7B models, for the sake of time efficiency. We have designed a comparative evaluation of model performances for two-stage and one-stage training strategies. The two-stage training referred to here is as described in Section~\ref{subsection:training}, which involves continuing training from the checkpoints of the first stage (the MuMath models). The one-stage training, directly applies the second stage of training on the base models. On both settings, we vary the data volumes of $\mathcal{D}_{\mu\text{-}code}$. Table~\ref{table:two_stage} illustrates the performance comparison of models derived from both strategies across different data volumes, revealing that training solely on $\mathcal{D}_{\mu\text{-}code}$ is worse than the two-stage training. Furthermore, by merging the training data from both stages into a single dataset for one-stage training, we observe that the outcomes are still not as favorable as those obtained from two separate training stages. 

To further validate the effectiveness of our two-stage training strategy, we select MetaMath~\citep{yu2023metamath} and Xwin-Math~\citep{xwin-math-github} 7B models as the initial checkpoints for Stage-2 training, emulating the scenario where relevant datasets were employed during the first stage (Given the the unavailability of the most recent models and dataset proposed in~\citep{xwin_math}, we opt to utilize Xwin-Math-7B-V1.0 detailed in the corresponding GitHub repository). Table~\ref{table:two_stage} illustrates that models fine-tuned from MetaMath and Xwin-Math checkpoints on $\mathcal{D}_{\mu\text{-}code}$ (two-stage) outperform the one directly trained from Llama (single-stage), verifying the efficacy of a two-stage training strategy as well as the compatibility of our $\mathcal{D}_{\mu\text{-}code}$ with different first-stage CoT datasets.

\subsection{Ablation Studies} 

\begin{table}
\huge
\centering
\resizebox{.48\textwidth}{!}{
    \begin{tabular}{ccccc}
    \toprule[2.2pt] 
    \multirow{2}{*}{Sythesized Solutions }& \multicolumn{2}{c}{LLaMA} & \multicolumn{2}{c}{CodeLlama} \\ 
    & GSM8K & MATH & GSM8K & MATH \\ 

    \midrule
    
    w all & 83.8 & 48.8 & 82.6 & 52.4 \\
    w/o prefix CoT & 81.3 & 47.5 & 81.8 & 49.4 \\ 
    w/o code debugging & 82 & 47.1 & 82.1 & 52.1 \\
    w/o either & 81.0 & 46.8 & 81.3 & 49.0\\

    \bottomrule[2.2pt] 
    \end{tabular}}
    \caption{Ablation study for prefix CoT and code debugging. }
    \label{table:ablation1}
\end{table}

\begin{table}
\huge
\centering
\resizebox{.48\textwidth}{!}{
    \begin{tabular}{cccccc}
    \toprule[2.2pt] 
    \multirow{2}{*}{Data Size} & \multirow{2}{*}{Pseudo-Answer} & \multicolumn{2}{c}{LLaMA} & \multicolumn{2}{c}{CodeLlama} \\ 
     & & GSM8K & MATH & GSM8K & MATH \\ 

    \midrule
    
    \multirow{2}{*}{30K} & w & 65.4 & 33.4 & 67.3 & 38.5 \\
     & w/o & 65.9 & 32.4 & 67.6 & 36.9 \\ 
     \midrule 
     \multirow{2}{*}{60K} & w & 71.4 & 37.6 & 73.4 & 42.7 \\ 
     & w/o & 71.2 & 36.6 & 72.3 & 40.4 \\ 
     \midrule 
     \multirow{2}{*}{90K} & w & 75.1 & 39.3 & 75.2 & 44.5 \\ 
     & w/o & 74.8 & 37.9 & 74.2 & 41.9 \\ 
     \midrule 
     \multirow{2}{*}{120K} & w & 76.1 & 40.7 & 76.9 & 45.7 \\ 
     & w/o & 74.6 & 40.3 & 75.9 & 43.6 \\ 
     \midrule 
     \multirow{2}{*}{150K} & w & 77.8 & 42.7 & 76.8 & 46 \\ 
     & w/o & 75.8 & 41.5 & 76.4 & 45 \\ 
     \midrule 
     \multirow{2}{*}{180K} & w & 77.3 & 43.5 & 78.5 & 47.3 \\ 
     & w/o & 76.7 & 42.7 & 77.7 & 46.5 \\ 

    \bottomrule[2.2pt] 
    \end{tabular}}
    \caption{Ablation study for pseudo-answer guidance filtering. }
    \label{table:ablation2}
\end{table}

\begin{figure*}[t] 
    \centering 
    \subfigure[Test on GSM8K]{\includegraphics[width=.48\linewidth]{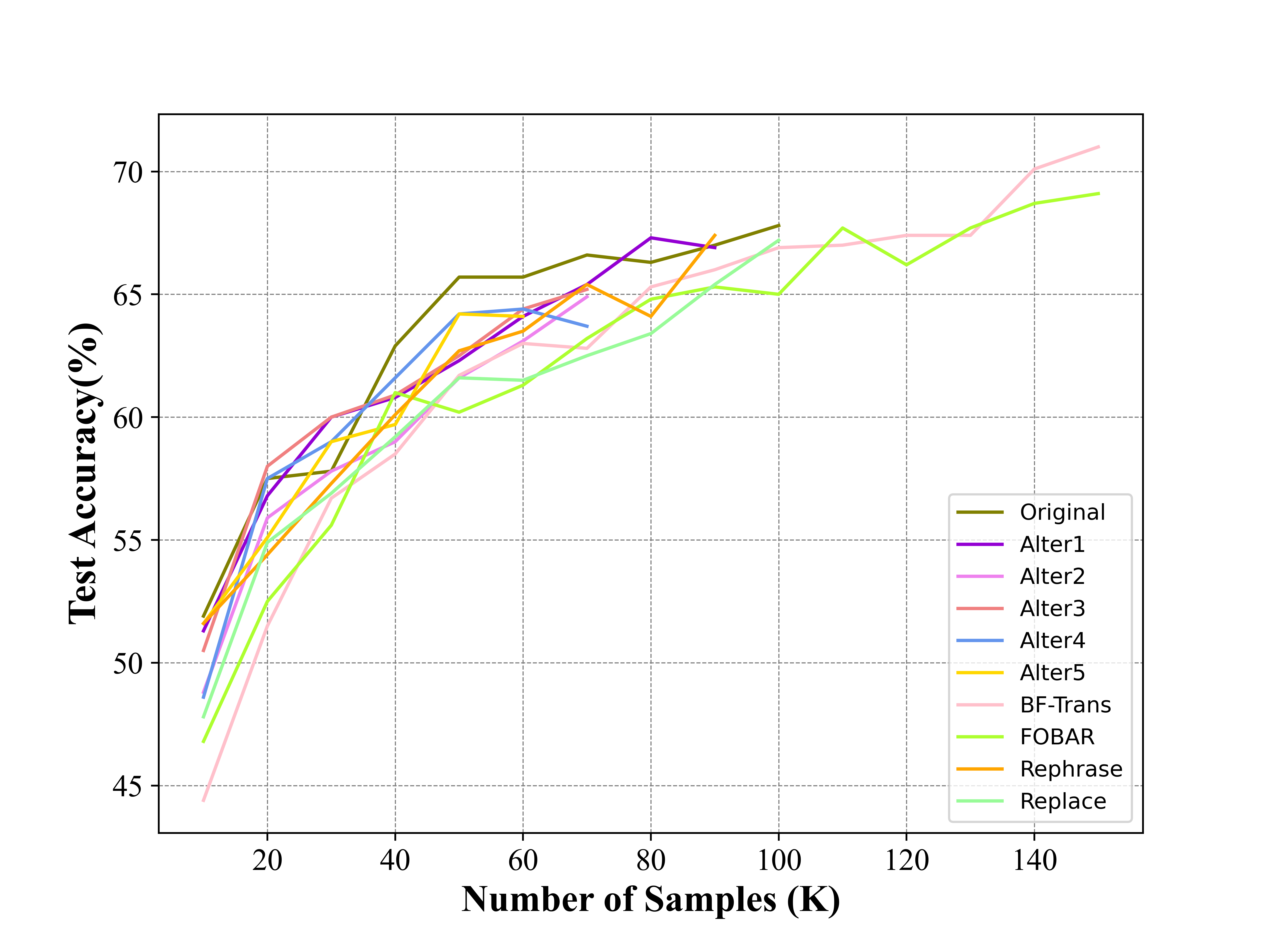}} \hspace{10pt}
    \subfigure[Test on MATH]{\includegraphics[width=.48\linewidth]{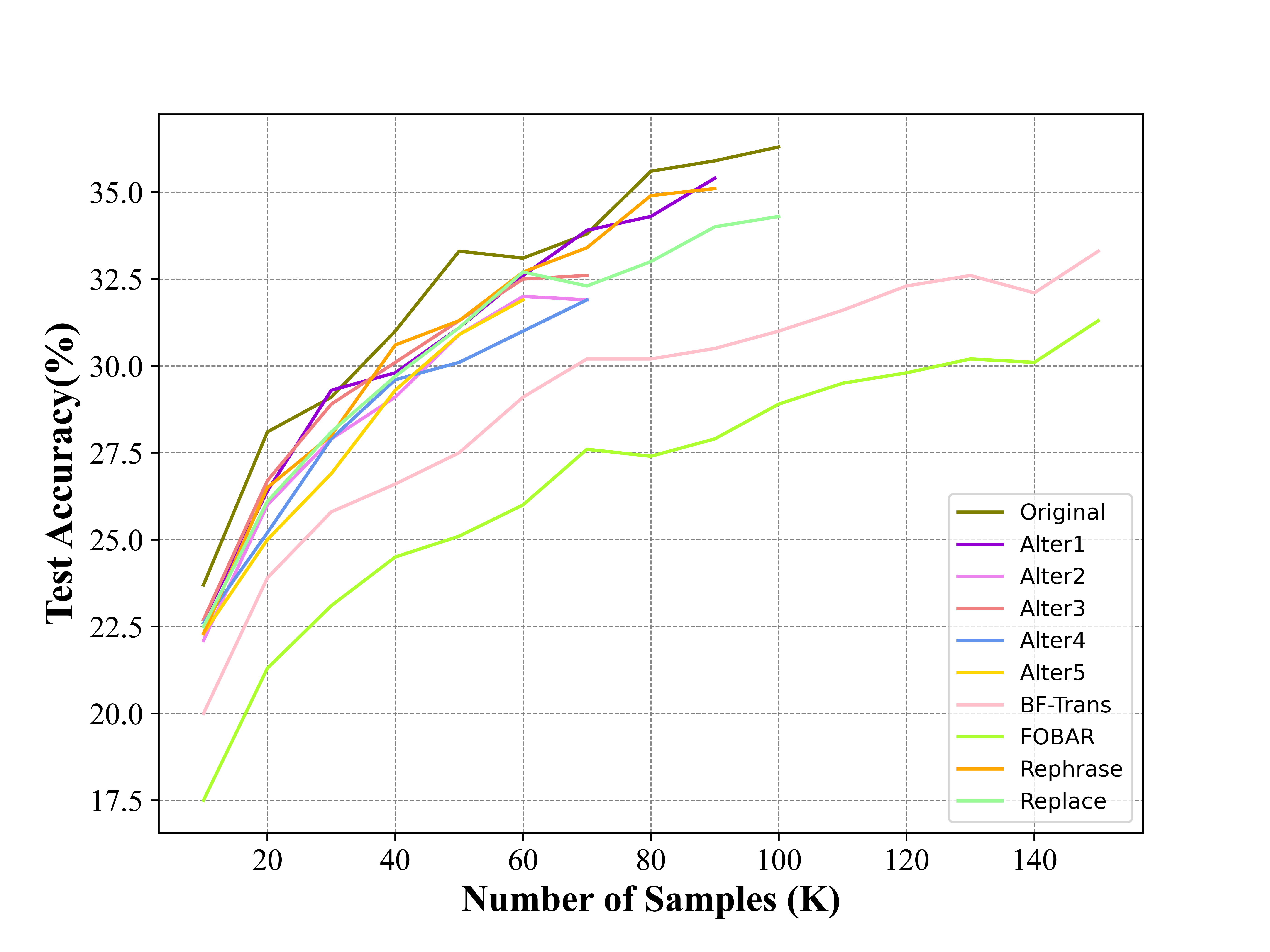}}
    \caption{Scaling all the subsets of MuMath-Code-Data. The models undergo a single stage (only Stage-2) of training. }
    \label{fig:scaling_single_stage} 
\end{figure*} 

To verify our proposed prefix CoT and code debugging, we respectively modify the solutions in $\mathcal{D}_{\mu\text{-}code}$ via two distinct approaches: the first approach involves the removal of the prefix CoT, thereby eliminating the detailed preliminary analysis and directly begining with code writing; the second approach consists of retaining only the final and successfully executed code and omitting all the other inexecutable code before as well as the corresponding debugging process. The results of this ablation study are presented in Table~\ref{table:ablation1}, which demonstrates that the exclusion of either the prefix CoT or code debugging leads to a decline in the models' test accuracy. This emphatically underscores the significance of a thorough analysis prior to code writing and the code mistake correction process for the models' learning.


Moreover, we conduct another ablation experiment on pseudo-answer guidance filtering. In Section~\ref{subsection:mumath_code_data}, we note that pseudo-answers are suitable for synthetic questions that lack a definitive correct answer, namely those in $Q_{alter}$ and $Q_{replace}$. In MuMath, majority voting is utilized to assign pseudo-answers to these questions. These pseudo-answers are then also employed to filter the data for $\mathcal{D}_{\mu\text{-}code}$ in the second training stage. As illustrated in Table~\ref{table:ablation2}, fine-tuning the model with data filtered through this pseudo-answer technique proves to be more beneficial than solutions obtained through directly random sampling. This trend holds across data volumes ranging from 30K to 180K.


\subsection{Scaling Study} 


The scaling experiments for various subsets of the MuMath-Code-Data are depicted in Figure~\ref{fig:scaling_single_stage}. These curves represent the performance changes of models trained on different data subsets with respect to the number of samples. The base model is LLaMA 7B and it is directly trained on the subsets of $\mathcal{D}_{\mu\text{-}code}$ (single-stage training). It is evident that with the increase in data volume, all subsets continuously contribute to the enhancement of the models' performance, and the curves still do not show saturation. This indicates that employing our methodology allows for the continued addition of data to further improve the LLMs' mathematical reasoning capabilities. For the two-stage scenario where the initial model is an intermediate checkpoint from Stage-1, please refer to Appendix~\ref{appendix:additional_experiments}.

\section{Conclusion} 
In this paper, we propose a multi-perspective and code integrated math reasoning dataset called MuMath-Code-Data, where each solution contains multi-turn code generation, code execution and pure natural language analysis (CoT). Through a two-stage training strategy, our MuMath-Code models outperforms the state-of-the-art open methods and even some powerful proprietary ones across different scales on the in-domain reasoning datasets (i.e., GSM8K and MATH) as well as those out-of-domain ones. Additionally, ablation studies demonstrates the effectiveness of our three novel methods for the data synthesis: prefix CoT, code debugging and pseudo-answer guidance filtering. Our work represents a new attempt at integrating mathematical question augmentation (tool-free) with code generation and execution (tool-use) to enhance the mathematical reasoning capabilities of LLMs, and we hope it can inspire subsequent research endeavors.

\section{Acknowledgments} 
This work was supported by National Key R\&D Program of China under Grant No. 2020AAA0104500, HY Project under Grant No. LZY2022033004, the Natural Science Foundation of Shandong Province under Grants No. ZR2020MF131 and No. ZR2021ZD19, the Science and Technology Program of Qingdao under Grant No. 21-1-4-ny-19-nsh, and Project of Associative Training of Ocean University of China under Grant No. 202265007.

\newpage 

\bibliography{custom}

\clearpage 

\appendix

\onecolumn

\section{Examples} 
\label{appendix:examples} 

\definecolor{blueviolet}{RGB}{138,43,226} 

\begin{exmp}{MuMath-Code-Data QA Pair}{exmp:qa}
\textbf{Question:} 

What is the smallest whole number that has a remainder of 1 when divided by 4, a remainder of 1 when divided by 3, and a remainder of 2 when divided by 5? \\

\textbf{Solution} 

\textcolor{violet}{We need to find a positive integer solution for the system of linear congruences. We can use the Chinese Remainder Theorem to solve this system. The Chinese Remainder Theorem states that for any system of equations like this, there exists a solution. Let's find the solution using sympy's solve\_congruence function.}\\

\textasciigrave{}\textasciigrave{}\textasciigrave{}python\\\\
from sympy.ntheory.modular import solve\_congruence \\
def smallest\_number():\\
\hspace*{5mm} congruences = [('1', '4'), ('1', '3'), ('2', '5')]\\ \hspace*{5mm} result = solve\_congruence(*congruences)\\
\hspace*{5mm} return  result[0]\\\\
result = smallest\_number()\\
print(result)\\\textasciigrave{}\textasciigrave{}\textasciigrave{}\\
\textasciigrave{}\textasciigrave{}\textasciigrave{}output\\
ValueError: 1 is not an integer\\
\textasciigrave{}\textasciigrave{}\textasciigrave{}\\
\textcolor{blueviolet}{The code above has encountered a problem. Now point out its mistakes and then correct them.}\\
Mistake:\\
The error is in the format of the input to the function `solve\_congruence`. The function is expecting tuples of the format `(remainder, modulus)`, but the inputs need to be integers, not strings.\\\\
Correction:\\
Let's modify the code to convert the strings to integers.\\\\
\textasciigrave{}\textasciigrave{}\textasciigrave{}python\\
def smallest\_number():\\
\hspace*{5mm} congruences = [(1, 4), (1, 3), (2, 5)]\\
\hspace*{5mm} result = solve\_congruence(*congruences)\\
\hspace*{5mm} return result[0]\\\\
result = smallest\_number()\\
print(result)\\
\textasciigrave{}\textasciigrave{}\textasciigrave{}\\
\textasciigrave{}\textasciigrave{}\textasciigrave{}output\\
NameError: name 'solve\_congruence' is not defined\\
\textasciigrave{}\textasciigrave{}\textasciigrave{}\\
\textcolor{blueviolet}{The code above has encountered a problem. Now point out its mistakes and then correct them.}\\
Mistake:\\
The error is that the `solve\_congruence` function is not defined. It should be imported from the `sympy` module before it can be used.\\\\
Correction:\\
Let's modify the code to import the `solve\_congruence` function from the `sympy` module.\\\\
\textasciigrave{}\textasciigrave{}\textasciigrave{}python\\
from sympy.ntheory.modular import solve\_congruence\\\\
def smallest\_number():\\
\hspace*{5mm} congruences = [(1, 4), (1, 3), (2, 5)]\\
\hspace*{5mm} result = solve\_congruence(*congruences)\\
\hspace*{5mm} return result[0]\\\\
result = smallest\_number()\\print(result)\\
\textasciigrave{}\textasciigrave{}\textasciigrave{}\\
\textasciigrave{}\textasciigrave{}\textasciigrave{}output\\
37\\
\textasciigrave{}\textasciigrave{}\textasciigrave{}\\
The smallest whole number that has a remainder of 1 when divided by 4, a remainder of 1 when divided by 3, and a remainder of 2 when divided by 5 is \$\textbackslash \textbackslash boxed\{37\}\$.
\end{exmp}

\vspace{1cm}

\section{Additional Experimental Results} 
\label{appendix:additional_experiments}

\begin{table}[h]
\centering
\setlength{\tabcolsep}{5.5mm} {
    
    \begin{tabular}{cccccc}
    \toprule
    \multirow{2}{*}{\makecell{$\mathcal{D}_{\mu\text{-}code}$ }} & \multirow{2}{*}{Training Strategy} & \multicolumn{2}{c}{LLaMA} & \multicolumn{2}{c}{CodeLlama} \\ 
     & & GSM8K & MATH & GSM8K & MATH \\ 

    \midrule
    
    \multirow{2}{*}{100K} 
     & $\mathcal{D}_{\mu\text{-}code} $ & 69.7 & 36.1 & 72.4 & 41 \\ 
     & $\mathcal{D}_\mu \rightarrow \mathcal{D}_{\mu\text{-}code} $ & 77.1 & 41.5 & 80.3 & 46.1 \\
     
     \midrule 
     
     \multirow{2}{*}{200K} 
     & $\mathcal{D}_{\mu\text{-}code}$ & 76.2 & 41.4 & 78.3 & 44.2 \\ 
     & $\mathcal{D}_\mu \rightarrow \mathcal{D}_{\mu\text{-}code} $ & 80.4 & 46 & 80.7 & 49.1 \\ 
     
     \midrule 
     
     \multirow{2}{*}{300K} 
     & $\mathcal{D}_{\mu\text{-}code}$ & 77.1 & 43.7 & 78.2 & 46.8 \\ 
     & $\mathcal{D}_\mu \rightarrow \mathcal{D}_{\mu\text{-}code} $ & 79.5 & 46.4 & 83.2 & 50.2 \\ 
     
     \midrule 
     
     \multirow{2}{*}{400K} 
     & $\mathcal{D}_{\mu\text{-}code}$ & 78 & 44.3 & 79 & 47.8 \\ 
     & $\mathcal{D}_\mu \rightarrow \mathcal{D}_{\mu\text{-}code} $ & 81.6 & 48.5 & 81.9 & 50.9 \\ 
     
     \midrule 
     
     \multirow{2}{*}{500K} 
     & $\mathcal{D}_{\mu\text{-}code}$ & 79.8 & 45.7 & 80.2 & 48.9 \\ 
     & $\mathcal{D}_\mu \rightarrow \mathcal{D}_{\mu\text{-}code} $ & 82.8 & 48.7 & 82.6 & 52.2 \\ 
     
     \midrule 
     
     \multirow{2}{*}{600K} 
     & $\mathcal{D}_{\mu\text{-}code}$ & 81.2 & 46.2 & 81 & 49.8 \\ 
     & $\mathcal{D}_\mu \rightarrow \mathcal{D}_{\mu\text{-}code} $ & 83.8 & 48.8 & 82.6 & 52.4 \\ 

    \bottomrule
    \end{tabular}
    }
    \caption{We vary the data volumes of $\mathcal{D}_{\mu\text{-}code}$. It is observed that training solely on $\mathcal{D}_{\mu\text{-}code}$ is consistently inferior to the two-stage training across all data volumes. }
    \label{table:two_stage_vary_sizes}
\end{table}

\clearpage 

\begin{figure}[H] 
    \centering 
    \subfigure[Test on GSM8K]{\includegraphics[width=.48\linewidth]{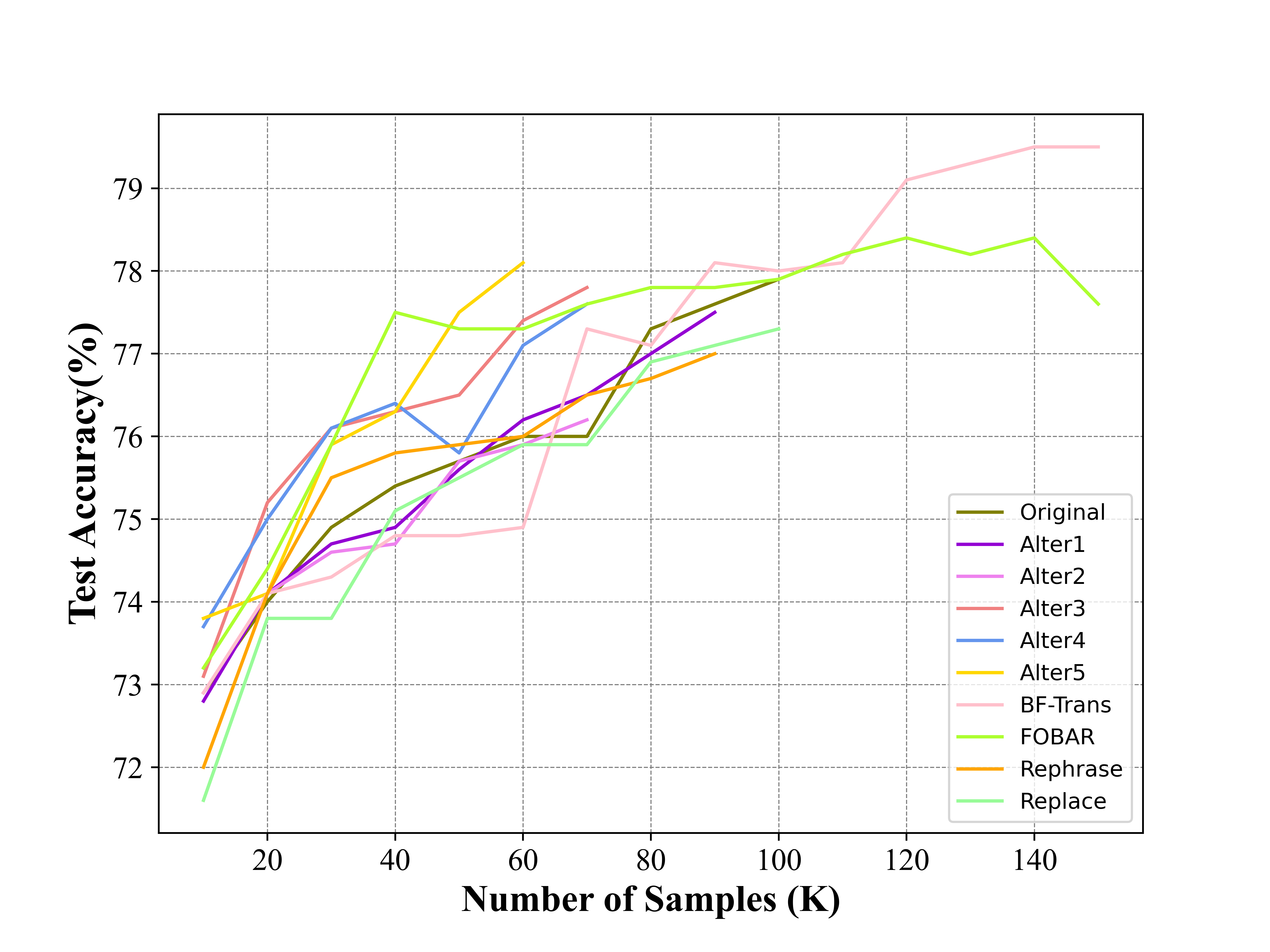}} \hspace{10pt}
    \subfigure[Test on MATH]{\includegraphics[width=.48\linewidth]{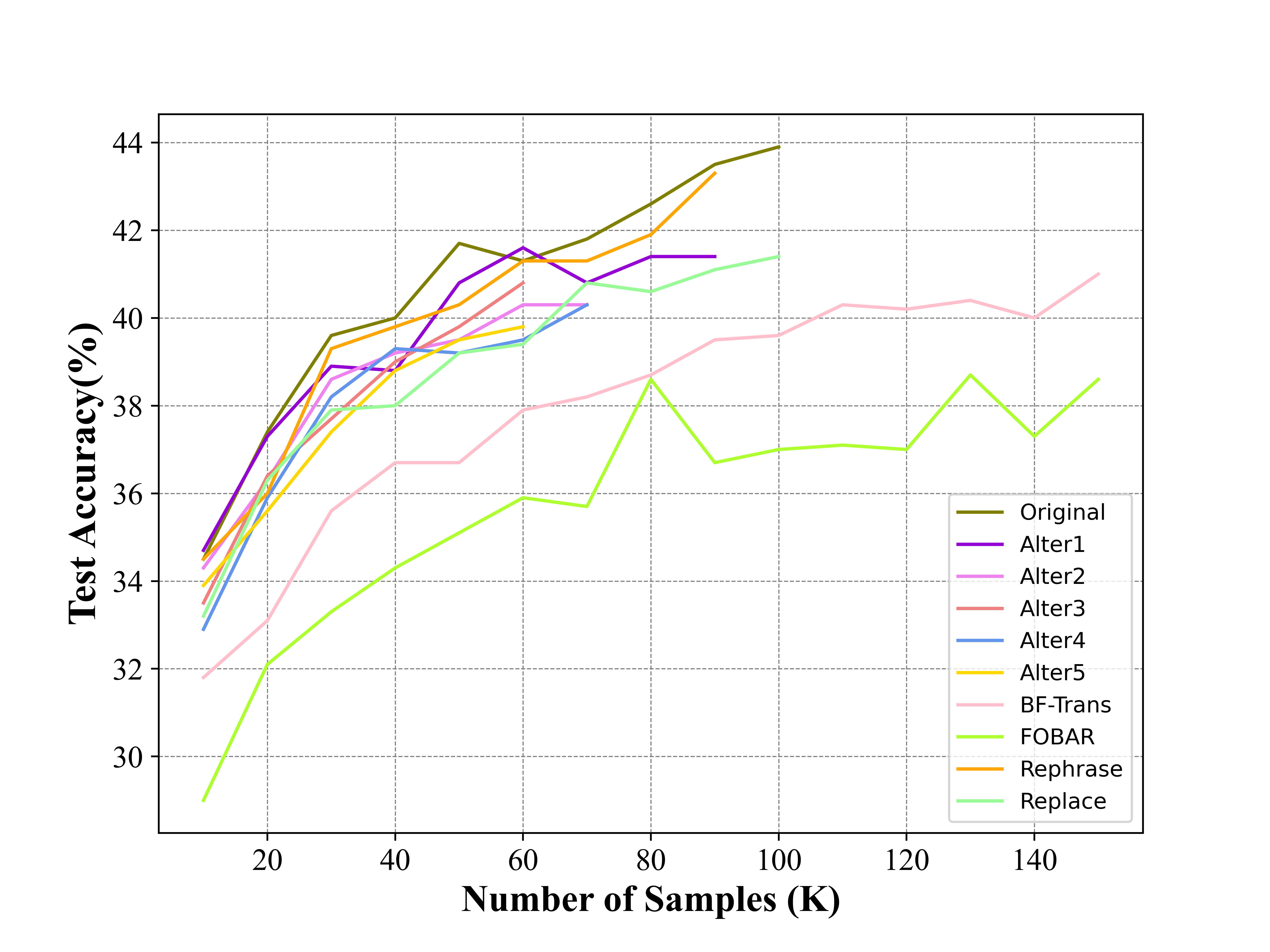}}
    \caption{Scaling all the subsets of MuMath-Code-Data. The model has already been finetuned on MuMath-Data. It is observable that the curves show very similar trends to those in Figure~\ref{fig:scaling_single_stage}.}
    \label{fig:scaling_two_stage} 
\end{figure} 

\twocolumn

\end{document}